\documentclass[sigconf]{acmart}
\definecolor{cvprblue}{rgb}{0.21,0.49,0.74}
\usepackage{booktabs}
\usepackage{multirow}

\usepackage{graphicx}
\usepackage{subcaption}
\usepackage{amsmath}
\usepackage[table]{xcolor} 
\usepackage{colortbl}      
\usepackage{pifont}
\newcommand{\cmark}{\ding{51}}
\usepackage{microtype}
\usepackage{balance}
\newcommand{\ie}{{\emph{i.e.}}, }

\newcommand{\eg}{{\emph{e.g.}}, }

\AtBeginDocument{%
  }

\copyrightyear{2026}
\acmYear{2026}
\setcopyright{cc}
\setcctype{by-nc-nd}
\acmConference[MM '26]{Proceedings of the 34th ACM International Conference on Multimedia}{November 10--14, 2026}{Rio de Janeiro, Brazil}
\acmBooktitle{Proceedings of the 34th ACM International Conference on Multimedia (MM '26), November 10--14, 2026, Rio de Janeiro, Brazil}
\acmDOI{10.1145/3767308.3834983}
\acmISBN{979-8-4007-2213-4/2026/11}

\begin{document}

\title{Seeing Semantic Shift: Difference-Aware Sentence-Level Temporal Segmentation of Sign Language Videos}

\author{Bowen Guo}
\affiliation{%
  \institution{State Key Laboratory for Novel Software Technology, Nanjing University}
  \city{Suzhou}
  \country{China}}
\email{bowen@smail.nju.edu.cn}

\author{Shiwei Gan}
\authornote{Corresponding authors.}
\affiliation{%
  \institution{State Key Laboratory for Novel Software Technology, Nanjing University}
  \city{Suzhou}
  \country{China}}
\email{sw@nju.edu.cn}

\author{Yafeng Yin}
\authornotemark[1]
\affiliation{%
  \institution{State Key Laboratory for Novel Software Technology, Nanjing University}
  \city{Suzhou}
  \country{China}}
\email{yafeng@nju.edu.cn}

\author{Xiao Liu}
\affiliation{%
  \institution{State Key Laboratory for Novel Software Technology, Nanjing University}
  \city{Suzhou}
  \country{China}}
\email{liuxiaox@smail.nju.edu.cn}

\author{Kuizhuang Liu}
\affiliation{%
  \institution{State Key Laboratory for Novel Software Technology, Nanjing University}
  \city{Suzhou}
  \country{China}}
\email{liukz@smail.nju.edu.cn}

\author{Zhiwei Jiang}
\affiliation{%
  \institution{State Key Laboratory for Novel Software Technology, Nanjing University}
  \city{Suzhou}
  \country{China}}
\email{jzw@nju.edu.cn}

\author{Lei Xie}
\affiliation{%
  \institution{State Key Laboratory for Novel Software Technology, Nanjing University}
  \city{Nanjing}
  \country{China}}
\email{lxie@nju.edu.cn}




\begin{abstract}
Recent advances in sign language understanding have achieved impressive success on short, single-sentence videos, yet their performance drops sharply when applied to long, continuous sign language videos. 
To bridge this gap, we focus on a challenging and realistic setting: Visual-only Sentence-level Sign Language Segmentation (Vis-SSLS), which aims to partition continuous sign language videos into non-overlapping sentence-level segments without any caption assistance, serving as a crucial prerequisite for downstream recognition and translation tasks.
However, sentence transitions in sign language are often smooth and visually ambiguous, lacking explicit pauses or posture resets. As a result, static frame representations may fail to capture the subtle temporal changes that indicate sentence boundaries.
To address this challenge, we propose \textbf{SignShift}, a difference-aware segmentation framework that explicitly models frame-to-frame feature variation as semantic cues for sentence boundary detection.
First, to model the feature variation, we design a Temporal Difference Module, which incorporates full-frame, facial, and hand cues, and employs inter-frame differencing to learn multi-scale temporal variations that capture both fine-grained local kinematics and global semantic transitions.
Second, to mitigate over- and under-segmentation issues, we design a Segment Count Prediction module, which predicts the number of sentences to guide boundary selection.
Extensive experiments on benchmark datasets demonstrate that SignShift substantially 
outperforms existing methods, validating its effectiveness. 
\end{abstract}
  
\maketitle
\section{Introduction}
Sign language research has made significant progress in recent years, primarily focusing on tasks such as sign language recognition (SLR)~\cite{zhang2023c2st,wei2023improving} and  sign language translation (SLT)~\cite{ye2024improving,yasser2024sign,shen2024auslan,gan2024signgraph}.
These tasks aim to bridge the communication gap between the deaf and hearing communities by converting sign language videos into gloss sequences or spoken-language sentences.
While these methods have achieved impressive results on short, single-sentence videos, their performance deteriorates when applied to long videos containing multiple sentences. 
Without explicit sentence boundaries, current models fail to maintain long-term visual-linguistic alignment, resulting in severe error accumulation across continuous signing streams.
This limitation significantly hinders the large-scale deployment of sign language technology, since natural signing communication typically consists of continuous streams rather than isolated sentences. 

\begin{figure}[t]
    \centering
    \includegraphics[width=0.48\textwidth]{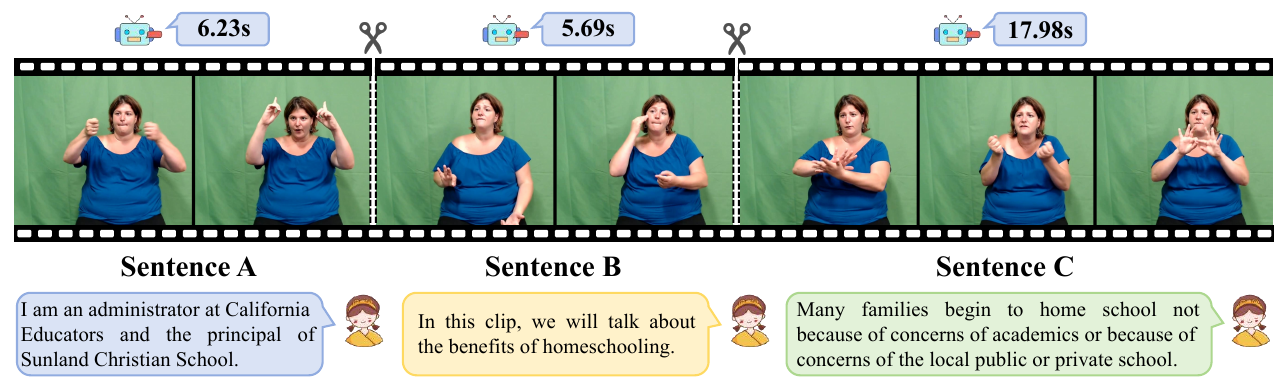}
    \caption{
        Overview of the Vis-SSLS task. 
        Given a long continuous signing video, the goal is to detect sentence boundaries and segment it into coherent sentence-level clips. 
        (Note: only videos are available, while captions are NOT available.)
    }
    \vspace{-8mm} 
    \label{fig:task}
\end{figure}

To bridge this gap, we study the problem of Sentence-level Sign Language Segmentation (SSLS), which is first proposed by Guo et al.~\cite{guo2025sentence}. 
Unlike previous works~\cite{guo2025sentence,bull2021aligning} that rely heavily on aligned captions to assist segmentation, we tackle a more rigorous and realistic setting: \textbf{Visual-only Sentence-level Sign Language Segmentation (Vis-SSLS)}.
As illustrated in Fig.~\ref{fig:task}, Vis-SSLS aims to partition continuous signing videos into non-overlapping sentence-level segments relying solely on visual cues.
Vis-SSLS is crucial for supporting downstream SLR and SLT on realistic long videos, and it also facilitates the construction of large-scale datasets by automatically generating sentence-level clips.
However, without the assistance of captions, Vis-SSLS is inherently more challenging than previous works~\cite{ding2018audio,stafylakis2018zero,albanie2020bsl,momeni2020watch}.
Due to the absence of dedicated architectures for Vis-SSLS, a straightforward solution might be adapting prevailing Temporal Action Segmentation (TAS) models (\eg MS-TCN~\cite{farha2019ms}, ASFormer~\cite{yi2021asformer}).
However, Vis-SSLS is fundamentally different from TAS. Specifically, TAS focuses on locating temporal boundaries between distinct action categories, where each temporal segment is typically associated with an action label.
Whereas Vis-SSLS aims to identify latent semantic transitions between sentences that appear visually continuous, it is a semantic segmentation problem, where each sentence corresponds to multiple actions. 
This distinction makes conventional action-centric segmentation paradigms inadequate for Vis-SSLS, as existing TAS approaches face three primary challenges:
First, sentence transitions in sign language are often smooth and lack explicit pauses, making boundaries visually ambiguous. Relying solely on static frame-level visual features fails to capture fine-grained temporal changes, which are essential for boundary localization.
Second, existing TAS models pay limited attention to local cues such as facial and hand movements, despite their critical importance for capturing subtle transitions between sentences and improving boundary accuracy.
Third, the absence of global sentence-count constraints leads to severe over- or under-segmentation.

Therefore, a paradigm shift from \emph{action-centric classification} to \emph{semantic boundary localization} is urgently required.
To address the aforementioned challenges, we propose a difference-aware segmentation framework tailored for Vis-SSLS. 
First, to overcome the limitations of static visual representations, we explicitly model temporal feature variations through multi-scale temporal differences, making the network sensitive to subtle transition dynamics.
Second, to capture the multi-articulator coordination inherent in sign language, we integrate fine-grained local kinematics from hand meshes and facial features into the global visual stream, enhancing the representation of discriminative local cues for sentence boundary detection.
Third, to prevent fragmented predictions, we introduce a Segment Count Prediction (SCP) module.
This module anticipates the total number of sentences in the video, injecting a global consistency prior. This prior constrains the predicted segmentation to align with the semantic structural distribution.
Our contributions are summarized as follows:
\begin{itemize}
    \item We introduce and formalize the task of Visual-only Sentence-Level Sign Language Segmentation, a crucial step for advancing sign language understanding in realistic long-video scenarios.
    \item We propose a difference-aware framework that rethinks boundary modeling from a temporal variation perspective. Specifically, our Temporal Difference Module extracts multi-scale temporal variations from global, facial, and hand features through temporal differencing. These difference-aware representations are then fused to capture both local kinematics and global semantic shifts for precise boundary detection.
    \item We design a Segment Count Prediction module that predicts the number of sentences in a video and imposes a global segment count on segmentation to mitigate over- and under-segmentation.
    \item We conduct extensive experiments on benchmark datasets, demonstrating that our approach achieves significantly more accurate segmentation compared with state-of-the-art baselines.
\end{itemize}

\section{Related Work}

\subsection{Alignment and Segmentation in Sign Language}
Existing research on sign language alignment and segmentation can be broadly categorized into gloss-level and sentence-level approaches based on semantic granularity.
Early efforts focused on gloss-level alignment, aiming to localize isolated signs or gloss boundaries using handcrafted visual features and probabilistic models, such as CRFs~\cite{yang2008sign,yang2006detecting}, HMMs~\cite{santemiz2009automatic}, and HSP trees~\cite{ong2014sign}. Subsequent works explored weakly supervised gloss alignment through coarse temporal supervision, including a priori mining~\cite{cooper2009learning} and multiple-instance learning~\cite{pfister2013large,alsolai2024automated}. Other methods incorporated auxiliary cues, such as mouthings~\cite{albanie2020bsl} and visual dictionaries~\cite{momeni2020watch}, to improve gloss localization. With the development of deep learning, CNN and transformer architectures~\cite{li2020transferring,varol2021read} further advanced frame-to-gloss alignment via attention mechanisms or sliding-window classifiers. Large-scale annotation pipelines also extended gloss alignment to unlabeled corpora by integrating visual keyword spotting~\cite{ding2018audio,stafylakis2018zero} and diarisation modules~\cite{albanie2021seehear}. However, these methods mainly focus on isolated gloss localization and struggle to capture sentence-level semantic structures in continuous signing.
Sentence-level approaches aim to identify boundaries between consecutive sentences. Skeleton-based methods~\cite{bull2020automatic} exploit pose dynamics to estimate potential transitions, while subtitle-guided approaches~\cite{bull2021aligning} utilize caption timestamps as temporal priors for segmentation. Recent methods, such as SignBD~\cite{guo2025sentence}, further incorporate caption-aware cues through textual-visual fusion to improve boundary precision.
Overall, existing methods either focus on gloss-level alignment or rely heavily on external textual supervision (\eg time-aligned subtitles) for sentence-level segmentation. This dependency limits their applicability in scenarios without precise annotations. In contrast, our work introduces a visual-only sentence-level segmentation paradigm by modeling multi-scale temporal differences and imposing a global segment-count constraint, enabling boundary detection solely from visual cues.

\begin{figure*}[t]
    \centering
    \includegraphics[width=\linewidth]{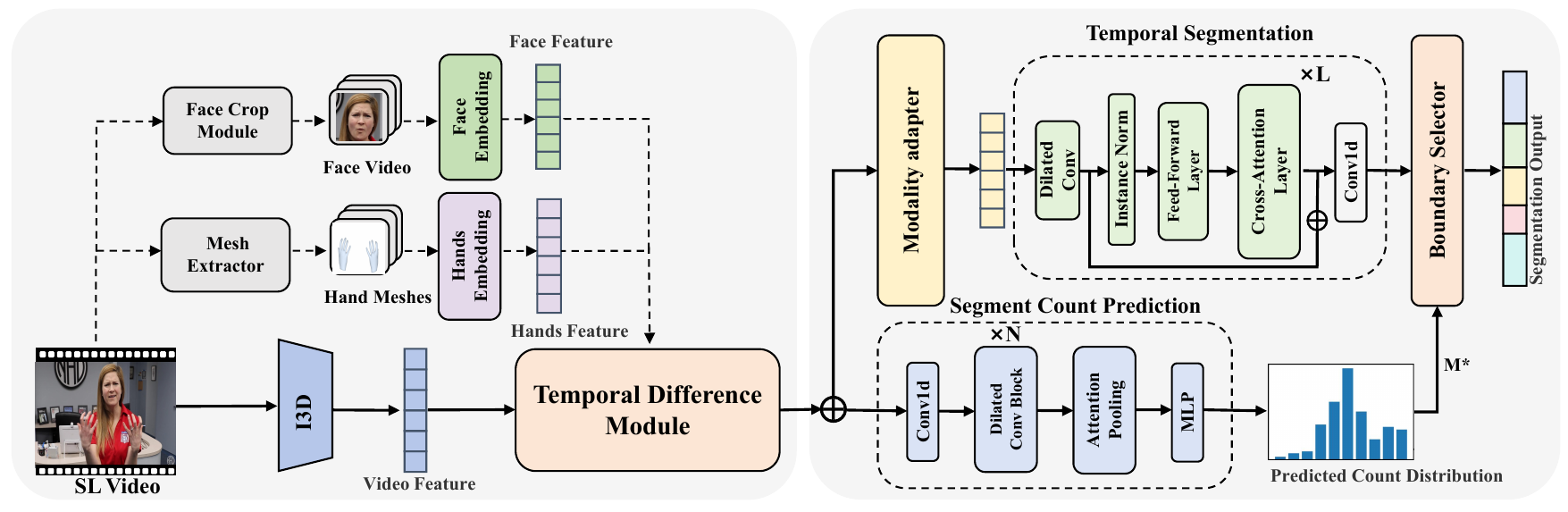}
    \caption{Overall framework of our proposed method.}
    \vspace{-4mm}
    \label{fig:framework}
\end{figure*}

\subsection{Temporal Action Segmentation}
Temporal Action Segmentation (TAS) is a video understanding task that typically aims to assign action labels to frames and identify temporal boundaries between consecutive actions. 
Existing studies have explored unsupervised~\cite{xu2024temporally,wang2022sscap,tran2024permutation,kumar2022unsupervised,shen2021learning}, weakly supervised~\cite{ghoddoosian2023weakly,lu2022set,rahaman2022generalized,sayed2023new,shen2022semi,fayyaz2020sct}, and fully supervised~\cite{shen2025understanding,lu2025multi,behrmann2022unified,lee2024error,liu2023diffusion,lu2024fact,pang2024long,shen2024progress} settings, with continuous improvements in temporal modeling and frame-wise action prediction.
Early approaches employed recurrent networks to capture temporal dependencies~\cite{singh2016multi}, followed by temporal convolutional networks (TCNs), such as MS-TCN~\cite{farha2019ms} and its variants~\cite{li2020ms}, which refined frame-level predictions through multi-stage temporal modeling. Transformer-based methods~\cite{yi2021asformer,wang2024cross,du2023dilated} further captured long-range dependencies via self-attention, while diffusion-based methods such as DiffAct~\cite{liu2023diffusion} formulated action sequence refinement as a generative denoising process. Recent methods further explored frame-action interaction~\cite{lu2024fact}, boundary uncertainty modeling~\cite{wang2020boundary}, segment-level relational reasoning~\cite{huang2020improving,zhang2022semantic2graph}, and temporal logic constraints~\cite{xu2022don} to improve segmentation consistency.
However, TAS and sign language segmentation differ fundamentally in boundary semantics. TAS focuses on action-level transitions, where boundaries are typically associated with changes in action categories or motion patterns. In contrast, sign language boundaries correspond to linguistic and semantic transitions, and a sentence-level segment may contain multiple actions. Therefore, existing TAS paradigms cannot be directly applied to Vis-SSLS, which requires modeling sentence-level semantic transitions rather than individual action changes.
To address this challenge, we propose a difference-aware segmentation framework that models multi-scale temporal variations as semantic boundary cues, together with a Segment Count Prediction module that provides a global sentence count prior.

\enlargethispage{\baselineskip}
\section{Method}
\subsection{Problem Setting}
Let a long sign language video be represented as a sequence of frame-level feature vectors 
$\mathbf{X} = [\mathbf{x}_1, \mathbf{x}_2, \dots, \mathbf{x}_T] \in \mathbb{R}^{T \times D}$, 
where $T$ denotes the total number of frames and $D$ is the feature dimension. 
The objective of Vis-SSLS is to identify a binary boundary sequence 
$\mathbf{y} = [y_1, y_2, \dots, y_T]$, 
where $y_t \in \{0, 1\}$ indicates whether frame $t$ is the ending boundary of a sentence ($y_t = 1$) 
or an internal frame ($y_t = 0$). 
Given the boundary sequence $\mathbf{y}$, 
we can derive an ordered set of boundary indices 
$\mathcal{T} = \{t_1, t_2, \dots, t_M\}$, 
where $0 < t_1 < t_2 < \dots < t_M = T$ are the exact frame indices satisfying $y_t = 1$. 
Crucially, $M$ denotes the total number of sentences in the video. 
The boundary set $\mathcal{T}$ partitions the continuous video into $M$ non-overlapping sentence-level segments 
$\mathcal{S} = \{S_1, S_2, \dots, S_M\}$, 
where each segment $S_i = [t_{i-1}+1, t_i]$ corresponds to one complete sign language sentence. 
Here, $t_0=0$ is a virtual index indicating that the first segment starts at frame 1; because boundaries are represented by sentence-ending frames, no additional boundary label is assigned to the left endpoint of the video. 
Formally, SignShift, denoted by $f_\theta(\cdot)$, learns to estimate the boundary probability for each frame, 
producing a continuous predictive sequence $\hat{\mathbf{y}} = f_\theta(\mathbf{X}) \in [0, 1]^T$, 
where $\hat{y}_t = P(y_t = 1 \mid \mathbf{X})$. 
During inference, the Boundary Selector uses the segment count predicted by SCP to select the corresponding number of internal boundaries from $\hat{\mathbf{y}}$ and fixes the final boundary at $T$, yielding the predicted boundary indices $\hat{\mathcal{T}}$ and sentence segments $\hat{\mathcal{S}}$.

\subsection{Overview of SignShift}
As illustrated in Figure~\ref{fig:framework}, SignShift performs Vis-SSLS in three stages. 
Given a continuous signing video, we first extract complementary representations from three visual streams, including global video features, facial features, and hand features. 
These features are fed into the Temporal Difference Module (TDM), which models frame-to-frame variations at multiple temporal scales to enhance cues related to sentence transitions.
The difference-enhanced representation is then processed by two parallel branches. 
The temporal segmentation branch predicts frame-wise boundary probabilities for locating candidate sentence boundaries. 
In parallel, the Segment Count Prediction (SCP) branch estimates the number of sentence-level segments in the input video, providing a global sentence count prior.
Finally, a Boundary Selector combines the frame-wise boundary probabilities with the predicted segment count to produce the final sentence-level segmentation results. 
The following sections describe each component in detail.

\subsection{Temporal Difference Module (TDM)}
\begin{figure}[t]
    \centering
    \begin{subfigure}[t]{0.45\textwidth}
        \centering
        \includegraphics[width=\textwidth]{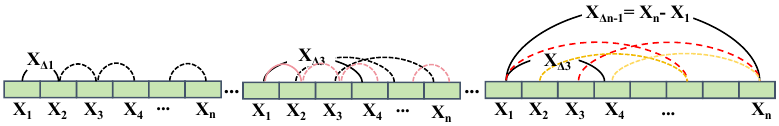}
        \caption{Multi-scale temporal difference.}
        \label{fig:tdm_a}
    \end{subfigure}
    \hfill
    \begin{subfigure}[t]{0.45\textwidth}
        \centering
        \includegraphics[width=\textwidth]{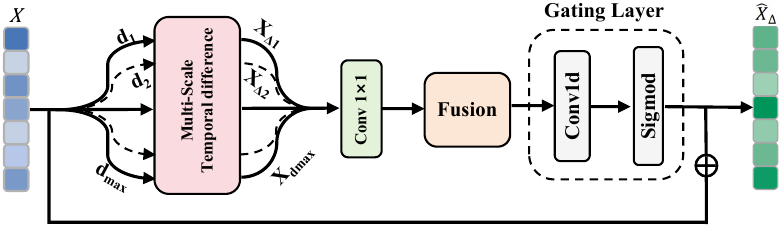}
        \caption{Overall structure of the Temporal Difference Module (TDM).}
        \label{fig:tdm_b}
    \end{subfigure}
    \caption{
    Illustration of the proposed TDM.
    (a) shows the multi-scale difference extraction process that captures temporal variations at different temporal offsets.
    (b) presents the overall TDM structure, which fuses the multi-scale difference features to produce enhanced representations.
    }
    \label{fig:tdm}
    \vspace{-6mm}  
\end{figure}

Continuous sign language videos usually exhibit smooth transitions between adjacent sentences, without explicit pauses or posture resets. 
As a result, sentence boundaries are often difficult to identify from static frame features alone. 
Instead, boundary-related cues are more likely to appear as subtle temporal variations, such as changes in hand motion or facial expression.
To capture such cues explicitly, we propose a Temporal Difference Module (TDM), illustrated in Figure~\ref{fig:tdm}, which models multi-scale frame-to-frame variations separately in the global video, hand, and facial feature streams and then fuses their difference-enhanced representations.

\noindent\textbf{Multi-scale temporal difference.}
Given an input video $V$, a visual backbone first extracts frame-level features $\mathbf{X} \in \mathbb{R}^{T \times D}$, where
\[
\mathbf{X} = [\mathbf{x}_1, \mathbf{x}_2, \dots, \mathbf{x}_T],
\]
and $\mathbf{x}_t \in \mathbb{R}^{D}$ denotes the feature at frame $t$.
As shown in Figure~\ref{fig:tdm_a}, TDM computes temporal differences under multiple offsets
$d \in \mathcal{D} = \{1,2,4,\dots,d_{\max}\}$, so as to capture temporal variations at different ranges.
For each offset $d$, the corresponding difference sequence $\mathbf{X}_{\Delta d} \in \mathbb{R}^{T \times D}$ is defined as
\begin{equation}
\mathbf{X}_{\Delta d}(t)=
\begin{cases}
\mathbf{x}_t - \mathbf{x}_{t-d}, & t > d,\\
\mathbf{0}, & t \le d,
\end{cases}
\qquad t = 1,\dots,T.
\label{eq:delta_d}
\end{equation}
Small offsets capture short-range temporal fluctuations, while larger offsets capture coarse-grained structural variations over longer temporal ranges.
The same multi-scale difference computation is applied independently to the global video, hand, and facial feature streams.
By integrating temporal variations from all three streams, TDM can adaptively localize boundaries regardless of whether a transition is an abrupt action shift or a gradual, subtle pause.

Following Figure~\ref{fig:tdm_b}, all difference sequences are first aligned to the same length by zero padding, and then projected into a shared latent space by a learnable $1\times1$ convolution. 
The resulting multi-scale difference feature is obtained by averaging over all offsets:
\begin{equation}
\mathbf{X}_{\Delta}
=
\frac{1}{|\mathcal{D}|}
\sum_{d \in \mathcal{D}}
\operatorname{Proj}(\mathbf{X}_{\Delta d}),
\label{eq:multi_scale_delta}
\end{equation}
where $\operatorname{Proj}(\cdot)$ denotes a shared $1\times1$ convolution.
To suppress noisy differences while preserving informative temporal changes, we further apply a learnable gating mechanism:
\begin{equation}
\mathbf{G} = \sigma\!\big(\operatorname{Conv1D}(\mathbf{X}_{\Delta})\big),
\label{eq:gate}
\end{equation}
\begin{equation}
\hat{\mathbf{X}}_{\Delta}
=
\mathbf{G} \odot \mathbf{X}_{\Delta}
+
\operatorname{Proj}(\mathbf{X}),
\label{eq:delta_hat}
\end{equation}
where $\sigma(\cdot)$ is the sigmoid function and $\odot$ denotes element-wise multiplication.
The gating term $\mathbf{G}$ adaptively emphasizes frames with salient temporal variation, while the residual projection of $\mathbf{X}$ preserves the original semantic context.

\noindent  \textbf{Local difference-aware fusion.}
In addition to the global motion information captured by the video stream,  
we incorporate local articulator information, \ie the hands and face,  
since they convey the most discriminative cues for sentence boundary transitions in sign language.
For the hands, we first obtain hand mesh sequences using HaMeR
\cite{pavlakos2024reconstructing},  
and then apply a ResNet-18 network to extract hand-mesh features $\mathbf{X}_h$.  
For the face, cropped facial regions are similarly fed into a ResNet-18 network to produce facial embeddings $\mathbf{X}_f$.
The same multi-scale temporal difference operation described above is applied independently to the hand and facial feature streams, producing their respective difference-enhanced representations:
\begin{equation}
\hat{\mathbf{X}}_{\Delta}^{h} = \operatorname{TDM}(\mathbf{X}_{h}), \quad
\hat{\mathbf{X}}_{\Delta}^{f} = \operatorname{TDM}(\mathbf{X}_{f}).
\label{eq:tdm_local}
\end{equation}
All difference-enhanced features are projected into a shared latent space and adaptively fused through learnable gating coefficients:
\begin{equation}
\hat{\mathbf{X}} =
\frac{1}{3} \big(
\operatorname{Proj}(\hat{\mathbf{X}}_{\Delta}) +
\alpha_h \operatorname{Proj}_h(\hat{\mathbf{X}}_{\Delta}^{h}) +
\alpha_f \operatorname{Proj}_f(\hat{\mathbf{X}}_{\Delta}^{f})
\big),
\label{eq:fusion}
\end{equation}
where $\operatorname{Proj}(\cdot)$, $\operatorname{Proj}_h(\cdot)$, and $\operatorname{Proj}_f(\cdot)$
denote learnable $1\times1$ convolutions that project the global (full-frame video), hand, and facial features,
and $\alpha_h$ and $\alpha_f$ are trainable scalar gates that control
the relative contribution of the hand and face streams.
The fused feature $\hat{\mathbf{X}}$ serves as the final difference-aware representation, capturing temporal feature variations by integrating global motion and fine-grained articulator cues.

\subsection{Segment Count Prediction (SCP)}
Although the TDM effectively captures multi-scale variations, it inherently struggles to distinguish between true inter-sentence boundaries and intra-sentence action transitions. Since a single sign language sentence typically contains multiple continuous signs, relying solely on difference-aware features makes the model vulnerable to local motion interference. Without a macroscopic view of the entire video, the model easily produces erroneous predictions, frequently manifesting as severe over-segmentation or under-segmentation.
To address this limitation, we introduce the Segment Count Prediction (SCP) module to estimate the probability distribution over the total number of sentences in a video. By formulating sentence count estimation as a global classification problem (\ie the $k$-th class means the video contains exactly $k$ sentences), SCP derives a holistic constraint that cannot be reliably inferred from local transitions. This predicted count acts as a powerful sentence count prior to regularize frame-level boundary decisions, ensuring structurally regulated and semantically consistent segmentation.

\noindent \textbf{Module Design.}
Given the fused difference-aware feature sequence $\hat{\mathbf{X}} = [\hat{\mathbf{x}}_1, \dots, \hat{\mathbf{x}}_T] \in \mathbb{R}^{T \times D}$, the SCP module aims to distill this dense, frame-level representation into a macroscopic structural prior. 
To capture hierarchical temporal dependencies, $\hat{\mathbf{X}}$ is first processed by a one-dimensional convolution and then through a stack of $N$ one-dimensional dilated convolutional layers. Let $\hat{\mathbf{X}}^{(0)}$ denote the output of the initial convolution. Formally, the $n$-th dilated layer operates as:
\begin{equation}
\hat{\mathbf{X}}^{(n)} = \phi\!\big(\operatorname{Conv1D}_n(\hat{\mathbf{X}}^{(n-1)})\big), \quad n = 1, \dots, N,
\label{eq:conv_layer}
\end{equation}
where $\phi(\cdot)$ denotes the ReLU activation, and $\operatorname{Conv1D}_n(\cdot)$ uses progressively increasing dilation rates to expand the temporal receptive field. The output $\hat{\mathbf{X}}^{(N)}$ captures temporal patterns at multiple scales, from local motion fluctuations to long-range semantic shifts.
Since not all frames contribute equally to the global sentence count estimation, we employ an attention-based temporal pooling mechanism (labeled as Attention Pooling in Figure~\ref{fig:framework}) to aggregate the sequence into a compact global descriptor. Specifically, a lightweight scoring network $\operatorname{Attn}(\cdot)$ evaluates the importance of each frame, which is then normalized to yield the attention weights $\boldsymbol{\alpha} = [\alpha_1, \dots, \alpha_T]^\top \in [0,1]^T$:
\begin{equation}
\boldsymbol{\alpha} = \operatorname{Softmax}\!\Big(\operatorname{Attn}\big(\hat{\mathbf{X}}^{(N)}\big)\Big).
\label{eq:global_alpha}
\end{equation}
The global embedding $\hat{\mathbf{x}}_{\text{g}} \in \mathbb{R}^{D}$ is subsequently obtained via a weighted summation of the temporal features:
\begin{equation}
\hat{\mathbf{x}}_{\text{g}} = \sum_{t=1}^{T} \alpha_t \, \hat{\mathbf{x}}^{(N)}_t.
\label{eq:global_pool}
\end{equation}
This attention mechanism empowers the model to dynamically balance the segmentation landscape. It simultaneously highlights subtle inter-sentence boundaries and suppresses intra-sentence motion interference.
Finally, the global embedding $\hat{\mathbf{x}}_{\text{g}}$ is mapped to a discrete probability distribution over all possible sentence counts using a Multilayer Perceptron (MLP):
\begin{equation}
\mathbf{p} = \operatorname{Softmax}\!\big(\mathrm{MLP}(\hat{\mathbf{x}}_{\text{g}})\big).
\label{eq:scp_pred}
\end{equation}
Here, $\mathbf{p} = [p_1, \dots, p_{M_{\text{max}}}]^\top$ represents the predicted count distribution, satisfying $\sum_{k=1}^{M_{\text{max}}} p_k = 1$.
The predefined hyperparameter $M_{\text{max}}$ denotes the maximum permissible limit of sentences, and each scalar $p_k$ represents the estimated probability that the video contains exactly $k$ semantic segments.

\noindent
\textbf{Loss function.}
To train the SCP module, we employ a Gaussian-smoothed cross-entropy loss that provides soft supervision over possible segment counts. 
Instead of using a one-hot target, we construct a continuous target distribution centered on the ground-truth count, which encourages smoother optimization and reduces sensitivity to near-correct predictions.
Let $\mathbf{p} = [p_1, p_2, \dots, p_{M_{\text{max}}}]$ denote the predicted probability distribution over possible segment counts, 
and let $M \in \{1, 2, \dots, M_{\text{max}}\}$ be the ground-truth number of segments. 
The smoothed target distribution $\mathbf{q}$ is defined by a normalized Gaussian kernel:
\begin{equation}
q_k = 
\frac{\exp\!\left(-\frac{(k - M)^2}{2\sigma^2}\right)}
{\sum_{j=1}^{M_{\text{max}}} \exp\!\left(-\frac{(j - M)^2}{2\sigma^2}\right)}, 
\quad k = 1, 2, \dots, M_{\text{max}},
\label{eq:gaussian_weight}
\end{equation}
where $\sigma$ controls the smoothness of the label distribution.
The classification loss is then formulated as the Kullback-Leibler divergence between the predicted distribution $\mathbf{p}$ and the smoothed target $\mathbf{q}$:
\begin{equation}
\mathcal{L}_{\text{SCP}} = 
\sum_{k=1}^{M_{\text{max}}} 
q_k \log \frac{q_k}{p_k}.
\label{eq:l_scp}
\end{equation}
This formulation allows small deviations around the true count to be softly penalized, improving stability in count prediction. 
The SCP module is trained independently using ground-truth sentence counts and is used only at inference to guide boundary selection.

\subsection{Boundary Learning and Selection}
\noindent
\textbf{Temporal Segmentation.}
The fused difference-aware feature sequence 
$\hat{\mathbf{X}}$ is first processed by a 
\emph{Modality Adapter} module (see Figure~\ref{fig:framework}) that consists of 
a series of feed-forward and self-attention layers.
This adapter refines the fused representation and enhances its temporal coherence, allowing the model to capture both fine-grained local cues and long-range dependencies.
We then adopt the widely used temporal segmentation backbone
ASFormer~\cite{yi2021asformer} for frame-wise sentence-boundary prediction.
Formally, the segmentation model 
$f_{\text{seg}}(\cdot)$ 
produces boundary confidence scores 
$\hat{\mathbf{y}} = f_{\text{seg}}(\hat{\mathbf{X}})$,
where
$\hat{\mathbf{y}} = [\hat{y}_1, \hat{y}_2, \dots, \hat{y}_T]$
and $\hat{y}_t$ denotes the predicted probability of frame $t$ being a valid sentence boundary.

\noindent
\textbf{Boundary Selection.}
To make the number of predicted segments consistent with the count estimated by SCP, we use the predicted segment count as a global prior during boundary selection.
Let $M^{*} = \arg\max_{k} p_k$ denote the most probable number of 
sentence segments.
Since $T$ is the index of the last video frame, we treat it as the fixed terminal boundary and select the remaining $M^{*}-1$ internal boundary indices based on the frame-wise confidence scores.
A temporal spacing constraint $\delta$ prevents boundaries from being placed too close to one another.
Formally, the optimal internal boundary set is obtained as
\begin{equation}
\mathcal{B}^{*} =
\arg\max_{\substack{
\mathcal{B} \subseteq \{1,\dots,T-1\},\\
|\mathcal{B}| = M^{*}-1,\\
|t_i-t_j| \ge \delta,\ \forall t_i \ne t_j \in \mathcal{B}\cup\{T\}
}}
\sum_{t \in \mathcal{B}} \hat{y}_t,
\label{eq:topk_boundary_refined}
\end{equation}
where $\mathcal{B}$ is any candidate set of internal boundary indices satisfying the spacing constraint.
This formulation effectively combines the frame-level boundary confidence with the global segment count constraint.
We then form the complete boundary set
$\mathcal{T}^{*}=\mathcal{B}^{*}\cup\{T\}=\{t_1<t_2<\cdots<t_{M^{*}}=T\}$,
which explicitly ensures that the final segment reaches the end of the video.
Given these refined boundary indices,
we construct the final segment set as
\begin{equation}
\hat{\mathcal{S}}
=
\left\{
(t_{i-1}, t_i]
\;\middle|\;
i = 1,\dots,M^{*},\;
t_0 = 0
\right\},
\label{eq:segment_construction}
\end{equation}
where each interval $(t_{i-1}, t_i]$ represents one predicted 
sentence-level segment.
This yields the final segmentation output $\hat{\mathcal{S}}$ in the form of 
a set of contiguous sentence-level segments.

\subsection{Segmentation Loss}
Following prior works on temporal segmentation~\cite{farha2019ms, li2020ms, yi2021asformer, lu2024fact}, 
we employ a combination of a Binary Cross-Entropy (BCE) loss and a mean squared error (MSE) for temporal smoothness
to supervise frame-wise boundary prediction. 
The total segmentation loss is expressed as
\begin{equation}
\mathcal{L}_{\text{seg}} = 
\mathcal{L}_{\text{bce}} + 
\lambda \, \mathcal{L}_{\text{smooth}},
\end{equation}
where $\lambda$ is a balancing hyperparameter that controls the contribution of the smoothness constraint (empirically set to $0.15$ in all experiments).
The BCE loss for boundary classification is given by
\begin{equation}
\mathcal{L}_{\text{bce}} = 
-\frac{1}{T}\sum_{t=1}^{T} 
\big[y_t \log(\hat{y}_t) + (1 - y_t)\log(1 - \hat{y}_t)\big],
\end{equation}
and the smooth loss is formulated as
\begin{equation}
\mathcal{L}_{\text{smooth}} = 
\frac{1}{T}\sum_{t=2}^{T} (\hat{y}_t - \hat{y}_{t-1})^2.
\end{equation}
Here, $y_t \in \{0,1\}$ indicates the true boundary label of frame $t$, 
and $\hat{y}_t \in [0,1]$ is the predicted boundary probability.

\begin{table*}[t]
\centering
\setlength{\tabcolsep}{3pt} 
\renewcommand{\arraystretch}{1.1}

\resizebox{\textwidth}{!}{%
\begin{tabular}{lccc|c|ccc|c|ccc|c|ccc|c}
\toprule
\multirow{3}{*}{\textbf{Model}} 
& \multicolumn{8}{c}{\textbf{How2Sign}} 
& \multicolumn{8}{c}{\textbf{OpenASL}} \\
\cmidrule(lr){2-9} \cmidrule(lr){10-17}

& \multicolumn{4}{c}{Dev} 
& \multicolumn{4}{c}{Test}
& \multicolumn{4}{c}{Dev} 
& \multicolumn{4}{c}{Test} \\
\cmidrule(lr){2-5} \cmidrule(lr){6-9} \cmidrule(lr){10-13} \cmidrule(lr){14-17}

& \multicolumn{3}{c|}{F1@\{10,25,50\}} & SER↓
& \multicolumn{3}{c|}{F1@\{10,25,50\}} & SER↓
& \multicolumn{3}{c|}{F1@\{10,25,50\}} & SER↓
& \multicolumn{3}{c|}{F1@\{10,25,50\}} & SER↓ \\
\midrule

MS-TCN~\cite{farha2019ms}      &64.93 /  &62.78 /  &45.12   &0.51   & 65.64 /  & 63.49 /  & 43.36  & 0.48 & 79.96 / &78.61 /  & 65.06  &0.29     &79.06 / &77.77 /  & 65.81 & 0.29   \\
MS-TCN++~\cite{li2020ms}    &69.29 /  &67.62 /  &48.03   &0.45  &72.91 / &70.16 / & 53.12  &  0.41 &76.23 /   &74.96 /   &62.22  &0.31   & 74.88 / &73.48 /  & 59.12  & 0.34   \\
ASFormer~\cite{yi2021asformer}    &72.86 /  &69.60 /  &49.44   &0.42 & 74.56 / & 72.93 /  & 53.80  & 0.36  &76.21 /  &74.87 /   &61.40  &0.33   & 74.50 / &  73.05 /& 58.88  & 0.35   \\
SignBD~\cite{guo2025sentence}     & 76.40 /  & 74.39 /  & 55.51  & 0.35  & 76.89 / & 73.97 / & 55.02  & 0.36  &79.23 /   & 78.42 /   & 64.99   & 0.29  &78.95 /   &  77.56 / & 65.03  & 0.30\\ 
DiffAct~\cite{liu2023diffusion}     &79.77 /  &76.43 /  &53.82   &0.33   &78.71 /  & 74.88 / & 55.69  &0.34   &80.25 /   &78.94 /   &65.83    &0.28   & 79.29 /  & 78.12 / & 65.93  &0.29    \\   
\rowcolor{gray!10} 
\textbf{SignShift}
& \textbf{83.30} /& \textbf{81.73} /& \textbf{61.43} & \textbf{0.23}
& \textbf{83.51} /& \textbf{81.61} /& \textbf{62.47} & \textbf{0.22}
& \textbf{84.72} / & \textbf{82.63} / & \textbf{67.87}  & \textbf{0.17}
& \textbf{84.79} /& \textbf{82.90} /& \textbf{68.25} & \textbf{0.18} \\
\bottomrule
\end{tabular}
}
\caption{
Comparison with state-of-the-art methods on How2Sign and OpenASL (Dev \& Test). 
 Best results are in \textbf{bold}.
}
\label{tab:main_results} 
\vspace{-6pt}
\end{table*}

\begin{table*}[t]
\centering
\normalsize
\setlength{\tabcolsep}{6pt}
\renewcommand{\arraystretch}{1.15}
\newcommand{\gc}[1]{\cellcolor{gray!10}#1}

\resizebox{\textwidth}{!}{
\begin{tabular}{c | c c c | ccc|c | ccc|c}
\toprule
\multirow{2}{*}{\textbf{Dataset}} 
& \multicolumn{3}{c|}{\textbf{Modules}} 
& \multicolumn{4}{c|}{\textbf{TS model: MS-TCN}} 
& \multicolumn{4}{c}{\textbf{TS model: ASFormer}} \\
\cmidrule(lr){2-4} \cmidrule(lr){5-8} \cmidrule(lr){9-12}
& TDM & Hands\&Face & SCP
& F1@10 & F1@25 & F1@50 & SER$\downarrow$
& F1@10 & F1@25 & F1@50 & SER$\downarrow$ \\
\midrule

\multirow{4}{*}{How2Sign~\cite{duarte2021how2sign}}
& -- & -- & --
& 64.93 & 62.78 & 45.12 & 0.51
& 72.86 & 69.60 & 49.44 & 0.42 \\
& \cmark & -- & --
& 68.83 & 66.33 & 47.60 & 0.45
& 75.10 & 73.77 & 55.23 & 0.39 \\
& \cmark & \cmark & --
& 71.02 & 69.69 & 51.71 & 0.43
& 77.99 & 75.15 & 53.40 & 0.35 \\
& \gc{\cmark} & \gc{\cmark} & \gc{\cmark}
& \gc{\textbf{81.05}} & \gc{\textbf{79.26}} & \gc{\textbf{61.82}} & \gc{\textbf{0.26}}
& \gc{\textbf{83.30}} & \gc{\textbf{81.73}} & \gc{\textbf{61.43}} & \gc{\textbf{0.23}} \\

\midrule

\multirow{4}{*}{OpenASL~\cite{shi2022open}}
& -- & -- & --
& 79.96 & 78.61 & 65.06 & 0.29
& 76.03 & 74.29 & 60.11 & 0.28 \\
& \cmark & -- & --
& 81.58 & 80.19 & 67.96 & 0.27
& 78.02 & 76.53 & 66.01 & 0.27 \\
& \cmark & \cmark & --
& 83.22 & 81.79 & 69.47 & 0.24
& 80.69 & 78.48 & 66.93 & 0.25 \\
& \gc{\cmark} & \gc{\cmark} & \gc{\cmark}
& \gc{\textbf{86.22}} & \gc{\textbf{84.48}} & \gc{\textbf{71.79}} & \gc{\textbf{0.17}}
& \gc{\textbf{84.72}} & \gc{\textbf{82.63}} & \gc{\textbf{67.87}} & \gc{\textbf{0.17}} \\

\bottomrule
\end{tabular}
}
\caption{
Ablation study of SignShift using different TAS models across datasets.
Best results are in \textbf{bold}.
}
\vspace{-6mm}
\label{tab:ablation_ts_models}
\end{table*}
\section{Experiments}

\subsection{Datasets}
We evaluate SignShift on two large-scale sign language datasets, 
\textbf{How2Sign}~\cite{duarte2021how2sign} and \textbf{OpenASL}~\cite{shi2022open}. 
Both datasets contain continuous signing videos aligned with sentence-level annotations. 
To adapt them to our Vis-SSLS task, 
we uniformly crop the raw videos into continuous clips of 0–10 minutes in duration, 
ensuring sufficient temporal context within each sample.
\textbf{How2Sign} is a multimodal and multiview corpus of continuous American Sign Language (ASL), 
consisting of over 80 hours of sign videos paired with speech, English transcripts, and depth information. 
We only use sign videos and sentence-level annotations in our experiments.
After preprocessing, we obtain 3,720 training, 461 development, and 464 test samples.
\textbf{OpenASL} is a large-scale ASL–English dataset collected from public online video platforms. 
It covers 288 hours of signing videos across multiple domains and includes over 200 different signers. 
Following the same preprocessing protocol, we obtain 11,023 training, 1,378 development, and 1,379 test samples.

\vspace{-4pt}
\subsection{Experimental Setup}
\textbf{Architecture Setting.}
All experiments are conducted on four RTX A6000 GPUs using PyTorch 1.13.
Video features (1024-D) are extracted with an I3D network pretrained on Kinetics. 
Hand meshes are obtained using HaMeR, and face regions are detected and cropped with BlazeFace. 
Both hand and face streams are encoded by pretrained ResNet-18 models into 512-D features.
(1) Temporal Difference Module: a stack of dilated 1D convolutions with maximum dilation $d_{\max}=16$.
(2) Segment Count Prediction Module: it consists of $N=4$ temporal convolutional layers and adopts Gaussian-smoothed count supervision with $\sigma=1.0$.
(3) Boundary Selection: the segmentation backbone follows the original ASFormer configuration~\cite{yi2021asformer}, and the minimum boundary spacing is set to $\delta=50$ during inference.

\noindent
\textbf{Training Setting.}
All models are trained for 60 epochs using the Adam optimizer with a weight decay of 0.0001. 
The initial learning rate is set to 0.0005 and the batch size to 4. 
The segmentation branch is trained with the classification loss and the smooth loss,
weighted by $\lambda = 0.15$ to balance their contributions.
The SCP module is trained independently using $\mathcal{L}_{\mathrm{SCP}}$ defined in Eq.~\eqref{eq:l_scp}, and its predicted segment count is used to guide boundary selection at inference time.
All baselines and SignShift share the same feature inputs, data splits, and evaluation protocol for fair comparison.

\noindent
\textbf{Evaluation Metrics.}
Following previous works~\cite{farha2019ms, yi2021asformer, liu2023diffusion}, 
we adopt segment-level evaluation metrics including F1@\{10, 25, 50\} and Segment Error Rate (SER). 
Specifically, the segmental F1 score evaluates the \textbf{segment-level} precision and recall based on Intersection over Union (IoU) at temporal overlap thresholds of 10\%, 25\%, and 50\%.
Meanwhile, SER measures the discrepancy between the predicted and ground-truth numbers of segments.


\vspace{-3mm}
\subsection{Overall Performance}
Table~\ref{tab:main_results} reports the overall segmentation results on the How2Sign and OpenASL benchmarks.
We compare SignShift with strong temporal segmentation baselines, including MS-TCN\cite{farha2019ms}, MS-TCN++\cite{li2020ms},  ASFormer\cite{yi2021asformer}, DiffAct\cite{liu2023diffusion}, and SignBD\cite{guo2025sentence}.
For a fair comparison, all baseline models are reimplemented and retrained under the same setting as SignShift. We report results on both the development and test sets. 

\noindent
\textbf{Evaluation on How2Sign Dataset.}
Specifically, compared with the strongest baseline DiffAct, SignShift achieves clear improvements on the key metric F1@50. 
On the development set, F1@50 increases from 53.82 to 61.43 (+7.61), while SER decreases from 0.33 to 0.23 (-0.10). 
On the test set, SignShift improves F1@50 from 55.69 to 62.47 (+6.78) and reduces SER from 0.34 to 0.22 (-0.12). 
These results demonstrate consistent gains over DiffAct.

\noindent
\textbf{Evaluation on OpenASL Dataset.}
SignShift achieves state-of-the-art performance on both the development and test sets. 
Compared with the strongest baseline DiffAct, it obtains higher F1@50 scores, increasing from 65.83 to 67.87 on the development set and from 65.93 to 68.25 on the test set. 
SER is also reduced notably, decreasing from 0.28 to 0.17 on the development set and from 0.29 to 0.18 on the test set, outperforming the previous SOTA model across all metrics.

\vspace{-2mm}
\subsection{Ablation Study}
We conduct extensive ablation studies to examine the contribution of each component in SignShift, 
including the TDM, local multi-articulator cues (Hands \& Face), and the SCP module. 
The experimental results are summarized in Table~\ref{tab:ablation_ts_models}.

\noindent
\textbf{Effect of TDM.}
We evaluate TDM by applying it to the global full-frame feature stream and comparing it with the corresponding baseline without temporal differencing. TDM consistently improves both backbones on both datasets. On How2Sign, it increases F1@50 from 45.12 to 47.60 with MS-TCN and from 49.44 to 55.23 with ASFormer, while reducing SER from 0.51 to 0.45 and from 0.42 to 0.39, respectively. On OpenASL, it improves F1@50 by 2.90 with MS-TCN and 5.90 with ASFormer, reducing SER to 0.27 for both. These results show that modeling multi-scale temporal differences in global video features facilitates the localization of subtle semantic transitions.

\noindent
\textbf{Effect of Hand and Face Cues.}
We further examine the contribution of integrating fine-grained
local kinematics, specifically hand meshes and facial regions.
Incorporating these local streams on top of the global visual TDM
consistently enhances boundary localization. On How2Sign, F1@10
increases from 68.83 to 71.02 with MS-TCN and from 75.10 to 77.99
with ASFormer. On OpenASL, it increases from 81.58 to 83.22 with
MS-TCN and from 78.02 to 80.69 with ASFormer. SER is also reduced
under both backbones on both datasets. These results indicate that
hand and facial cues provide discriminative information complementary
to macroscopic body movements for identifying sign language sentence
boundaries.

\noindent
\textbf{Effect of Backbone.}
We compare MS-TCN and ASFormer using the complete SignShift
configuration with TDM, Hands\&Face, and SCP, corresponding to the
last row of each dataset block in Table~\ref{tab:ablation_ts_models}.
On How2Sign, the ASFormer-based model achieves an F1@10 of 83.30 and
an SER of 0.23, compared with 81.05 and 0.26, respectively, for the
MS-TCN-based model. On OpenASL, the MS-TCN-based model achieves a
higher F1@50 of 71.79 than the 67.87 obtained with ASFormer, while
both models obtain an SER of 0.17. These results show that SignShift
remains effective with both convolutional and Transformer-based
backbones.

\noindent
\textbf{Effect of SCP on Segmentation.}
We analyze how the SCP module affects fine-grained boundary accuracy.
As reported in Table~\ref{tab:ablation_ts_models}, introducing SCP consistently improves F1@50 and reduces SER under both backbones.
With MS-TCN, F1@50 increases from 51.71 to 61.82 on How2Sign, with SER decreasing from 0.43 to 0.26.
On OpenASL, F1@50 increases from 69.47 to 71.79, while SER decreases from 0.24 to 0.17.
Similar improvements are observed with ASFormer.
These improvements indicate that the global segment-count prior provided by SCP helps regularize the predicted boundary distribution.

\begin{figure}[t]
    \centering
    \begin{minipage}[t]{0.48\linewidth}
        \centering
        \includegraphics[width=\textwidth]{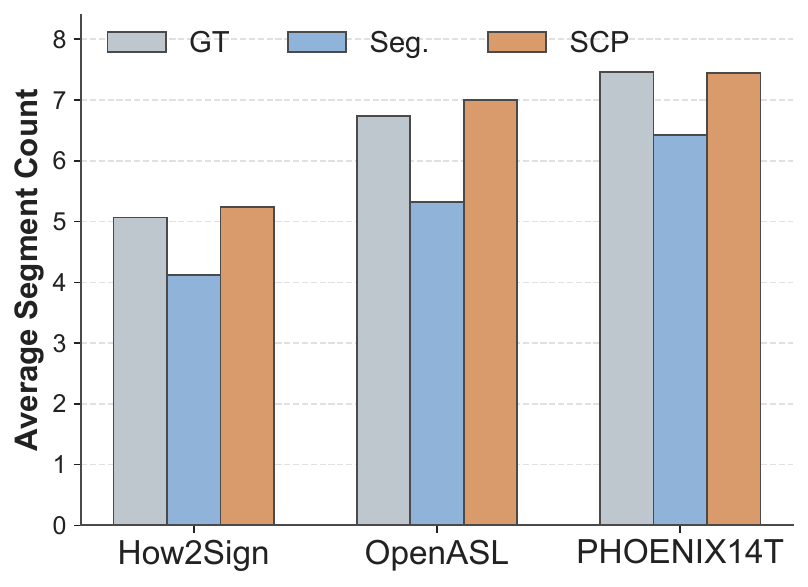}
        \subcaption{Average Segment Count} \label{fig:count_a}
    \end{minipage}
    \hfill
    \begin{minipage}[t]{0.48\linewidth}
        \centering
        \includegraphics[width=\textwidth]{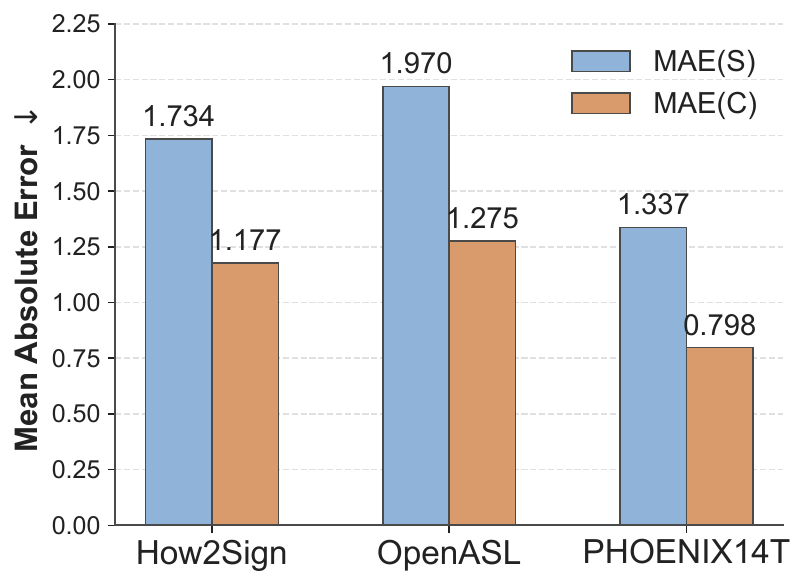}
        \subcaption{Count Estimation Error} \label{fig:count_b}
    \end{minipage}
    \caption{Segment count estimation results. (a) Comparison of the average segment count. (b) Mean Absolute Error (MAE) of count estimation. }
    \label{fig:segcount_analysis}
\end{figure}

\noindent
\textbf{Effect of SCP Accuracy.}
Figure~\ref{fig:segcount_analysis} presents the quantitative results
of segment count estimation across three datasets.
We denote the MAE of sentence counts inferred from the segmentation
output as \textit{MAE(S)}, and the MAE of counts directly predicted
by SCP as \textit{MAE(C)}, where S and C refer to segmentation and
count prediction, respectively. Compared with the
segmentation-derived counts, SCP reduces the MAE from 1.734 to 1.177
on How2Sign, from 1.970 to 1.275 on OpenASL, and from 1.337 to 0.798
on PHOENIX14T.
These results confirm that the SCP effectively models the global
structural patterns of signing videos, providing valuable guidance
for boundary calibration.

\begin{table}[t]
\centering
\small
\setlength{\tabcolsep}{8pt} 
\renewcommand{\arraystretch}{1.2}
\begin{tabular}{@{} c c ccc c @{}} 
\toprule
\multirow{2}{*}{\textbf{Duration (s)}} & \multirow{2}{*}{\textbf{Avg Frames}} & \multicolumn{3}{c}{\textbf{F1 Score (\%)}} & \multirow{2}{*}{\textbf{SER $\downarrow$}} \\
\cmidrule(lr){3-5} 
& & \textbf{@10} & \textbf{@25} & \textbf{@50} & \\
\midrule
17.9  & 449.8  & 85.19 & 83.55 & 66.80 & 0.25 \\
31.9  & 814.9  & 84.11 & 81.83 & 65.95 & 0.22 \\
78.71 & 1887.8 & 83.65 & 78.24 & 51.03 & 0.13 \\
\bottomrule
\end{tabular}
\caption{Segmentation results on videos of different durations on How2Sign dataset.}
\vspace{-7mm}
\label{tab:video_duration}
\end{table}

\noindent
\textbf{Effect of Video Length.}
We further examine the segmentation performance across videos of different durations, as reported in Table~\ref{tab:video_duration}.
On How2Sign, F1@10 remains relatively stable, decreasing from 85.19 on short clips to 83.65 on long clips.
In contrast, the stricter F1@50 decreases from 66.80 to 51.03, indicating that precise boundary localization becomes more difficult as video duration increases.
Meanwhile, SER decreases from 0.25 to 0.13, suggesting that the count-guided selector remains effective in limiting segmentation-count errors on longer videos.


\subsection{Computational Efficiency}
Table~\ref{tab:efficiency} compares the computational efficiency and segmentation performance of SignShift against representative baselines.
Lightweight models such as MS-TCN++ achieve lower computational costs but obtain limited segmentation performance. 
ASFormer improves segmentation accuracy through stronger temporal modeling, but introduces higher inference latency. 
Compared with ASFormer, SignShift achieves a substantial F1@50 improvement of 8.67 percentage points while maintaining comparable computational complexity and faster inference speed. 
Compared with SignBD and DiffAct, SignShift further improves boundary localization accuracy with competitive costs.


\begin{table}[t]
\centering
\small
\setlength{\tabcolsep}{4pt} 
\renewcommand{\arraystretch}{1.1}
\resizebox{\linewidth}{!}{
\begin{tabular}{l ccc c}
\toprule
\multirow{2}{*}{\textbf{Model}} & \multicolumn{3}{c}{\textbf{Efficiency}} & \textbf{Performance} \\
\cmidrule(lr){2-4} \cmidrule(lr){5-5}
& Params $\downarrow$ & FLOPs $\downarrow$ & Time $\downarrow$ & F1@50 $\uparrow$ \\
\midrule
MS-TCN++     & \textbf{0.89} & \textbf{1.42} & \textbf{18.61} & 43.36 \\
ASFormer     & 1.90 & 2.93 & 214.45 & 53.80 \\
SignBD       & 6.47 & 2.35 & 123.10 & 55.02 \\
DiffAct      & 2.85 & 4.94 & 65.96  & 55.69 \\
\midrule
\rowcolor{gray!10} 
SignShift    & 5.26 & 4.25 & 120.27 & \textbf{62.47} \\
\bottomrule
\end{tabular}
} 
\caption{Comparison of models in terms of Parameters (M), FLOPs (G), Inference Time (ms), and F1@50.}
\vspace{-8mm}
\label{tab:efficiency}
\end{table}

\subsection{Qualitative Analysis}
\begin{figure}[t]
    \centering
    \includegraphics[width=0.9\linewidth]{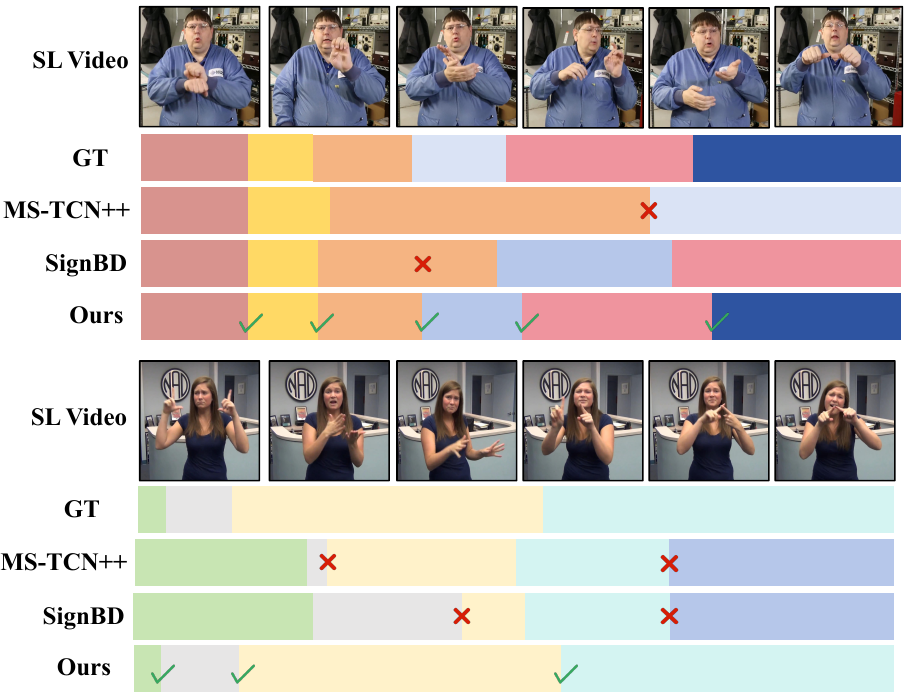}
    \caption{
    Qualitative results on OpenASL. The checkmarks ($\checkmark$) indicate correctly identified sentence boundaries by SignShift, whereas the crosses ($\times$) denote incorrect or missed boundaries predicted by the baseline models.
    }
    \label{fig:qualitative}
\end{figure}

Figure~\ref{fig:qualitative} presents qualitative comparisons on OpenASL against MS-TCN++ and SignBD. 
While MS-TCN++ tends to generate fragmented boundaries and SignBD yields smoother but misaligned segments, SignShift produces more consistent and semantically coherent results. 
These examples clearly demonstrate that modeling frame-to-frame differences enables more accurate boundary localization in long, continuous sign videos.

\begin{figure}[t]
    \centering
    \includegraphics[width=\linewidth]{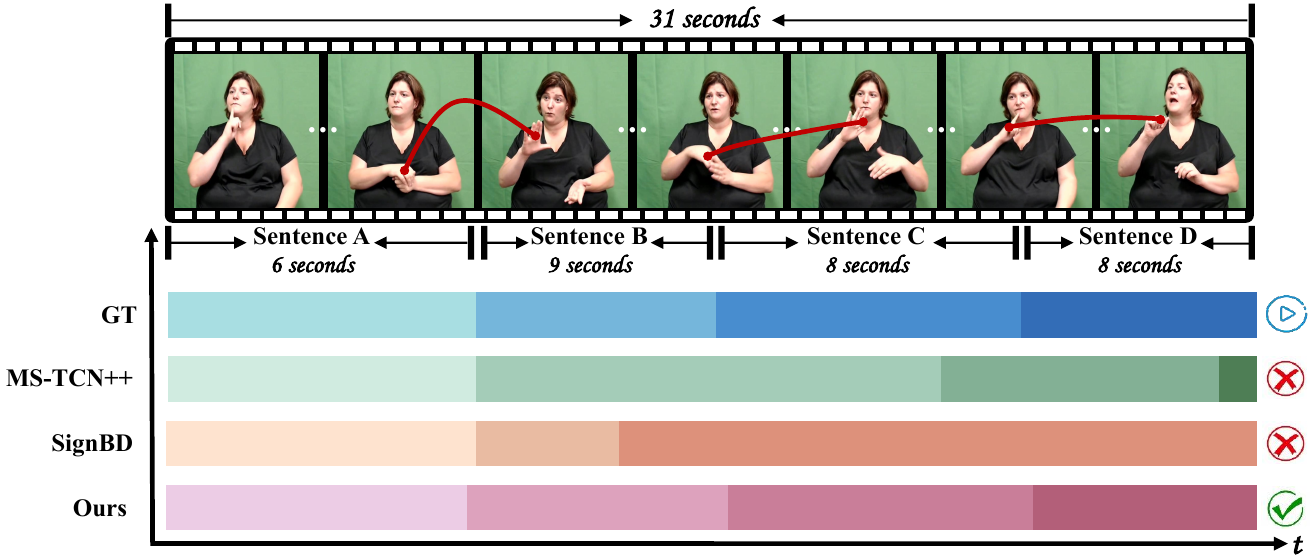} 
    \caption{Qualitative comparison of boundary detection.}
    \label{fig:qualitative_vis}
    \vspace{-4mm}
\end{figure}

Figure~\ref{fig:qualitative_vis} illustrates the challenge of visual continuity through a continuous signing example and its corresponding motion trajectory. While all models reliably detect obvious pauses, MS-TCN++ and SignBD struggle during fluid signing; MS-TCN++ predicts misaligned boundaries, whereas SignBD misses transitions and leads to under-segmentation.
Conversely, by leveraging TDM for fine-grained temporal modeling and SCP for global count constraints, SignShift better localizes subtle semantic boundaries, achieving closer alignment with the ground truth.

\section{Conclusion}
In this work, we focused on segmenting long continuous signing videos into coherent sentence-level clips using purely visual cues.
To address smooth transitions and ambiguous sentence boundaries, we proposed SignShift, a difference-aware segmentation framework that first models multi-scale temporal variations separately in global full-frame, hand, and facial features and then fuses the resulting difference-aware representations to enhance boundary perception.
Furthermore, the SCP module estimates the video-level sentence count and uses it as a global prior to guide boundary selection.
Experiments demonstrate that our framework achieves good segmentation results.

\begin{acks}
This work is supported in part by National Natural Science Foundation of China under Grant Nos. 62172208, 92467202, 62272216; Key Projects of Jiangsu Provincial Basic Research Program under Grant No. BK20243040; JiangSu Natural Science Foundation under Grant No. BK20251989. This work is partially supported by Fundamental and Interdisciplinary Disciplines Breakthrough Plan of the Ministry of Education of China (No. JYB2025XDXM118); the ``111 Center'' (No. B26023); Collaborative Innovation Center of Novel Software Technology and Industrialization.
\end{acks}

\bibliographystyle{ACM-Reference-Format}
\bibliography{bib/sls}

@inproceedings{bull2020automatic,
  title={Automatic segmentation of sign language into subtitle-units},
  author={Bull, Hannah and Gouiff{\`e}s, Mich{\`e}le and Braffort, Annelies},
  booktitle={European Conference on Computer Vision},
  pages={186--198},
  year={2020},
  organization={Springer}
}

@inproceedings{bull2021aligning,
  title={Aligning subtitles in sign language videos},
  author={Bull, Hannah and Afouras, Triantafyllos and Varol, G{\"u}l and Albanie, Samuel and Momeni, Liliane and Zisserman, Andrew},
  booktitle={Proceedings of the IEEE/CVF International Conference on Computer Vision},
  pages={11552--11561},
  year={2021}
}

@inproceedings{guo2025sentence,
  title={Sentence-level Segmentation for Long Sign Language Videos with Captions},
  author={Guo, Bowen and Gan, Shiwei and Yin, Yafeng and Liu, Xiao and Jiang, Zhiwei and Meng, Shunmei},
  booktitle={Proceedings of the 33rd ACM International Conference on Multimedia},
  pages={3428--3437},
  year={2025}
}

@inproceedings{santemiz2009automatic,
  title={Automatic sign segmentation from continuous signing via multiple sequence alignment},
  author={Santemiz, Pinar and Aran, Oya and Saraclar, Murat and Akarun, Lale},
  booktitle={2009 IEEE 12th International Conference on Computer Vision Workshops, ICCV Workshops},
  pages={2001--2008},
  year={2009},
  organization={IEEE}
}

@inproceedings{ong2014sign,
  title={Sign spotting using hierarchical sequential patterns with temporal intervals},
  author={Ong, Eng-Jon and Koller, Oscar and Pugeault, Nicolas and Bowden, Richard},
  booktitle={Proceedings of the IEEE conference on computer vision and pattern recognition},
  pages={1923--1930},
  year={2014}
}

@article{yang2008sign,
  title={Sign language spotting with a threshold model based on conditional random fields},
  author={Yang, Hee-Deok and Sclaroff, Stan and Lee, Seong-Whan},
  journal={IEEE transactions on pattern analysis and machine intelligence},
  volume={31},
  number={7},
  pages={1264--1277},
  year={2008},
  publisher={IEEE}
}

@inproceedings{yang2006detecting,
  title={Detecting coarticulation in sign language using conditional random fields},
  author={Yang, Ruiduo and Sarkar, Sudeep},
  booktitle={18th International conference on pattern recognition (ICPR'06)},
  volume={2},
  pages={108--112},
  year={2006},
  organization={IEEE}
}

@inproceedings{cooper2009learning,
  title={Learning signs from subtitles: A weakly supervised approach to sign language recognition},
  author={Cooper, Helen and Bowden, Richard},
  booktitle={2009 IEEE conference on computer vision and pattern recognition},
  pages={2568--2574},
  year={2009},
  organization={IEEE}
}

@article{pfister2013large,
  title={Large-scale learning of sign language by watching TV},
  author={Pfister, Tomas and Charles, James and Zisserman, Andrew},
  year={2013},
  publisher={British Machine Vision Association and Society for Pattern Recognition}
}

@article{alsolai2024automated,
  title={Automated sign language detection and classification using reptile search algorithm with hybrid deep learning},
  author={Alsolai, Hadeel and Alsolai, Leen and Al-Wesabi, Fahd N and Othman, Mahmoud and Rizwanullah, Mohammed and Abdelmageed, Amgad Atta},
  journal={Heliyon},
  volume={10},
  number={1},
  year={2024},
  publisher={Elsevier}
}

@inproceedings{albanie2020bsl,
  title={BSL-1K: Scaling up co-articulated sign language recognition using mouthing cues},
  author={Albanie, Samuel and Varol, G{\"u}l and Momeni, Liliane and Afouras, Triantafyllos and Chung, Joon Son and Fox, Neil and Zisserman, Andrew},
  booktitle={European conference on computer vision},
  pages={35--53},
  year={2020},
  organization={Springer}
}

@inproceedings{momeni2020watch,
  title={Watch, read and lookup: learning to spot signs from multiple supervisors},
  author={Momeni, Liliane and Varol, Gul and Albanie, Samuel and Afouras, Triantafyllos and Zisserman, Andrew},
  booktitle={Proceedings of the Asian Conference on Computer Vision},
  year={2020}
}

@inproceedings{li2020transferring,
  title={Transferring cross-domain knowledge for video sign language recognition},
  author={Li, Dongxu and Yu, Xin and Xu, Chenchen and Petersson, Lars and Li, Hongdong},
  booktitle={Proceedings of the IEEE/CVF conference on computer vision and pattern recognition},
  pages={6205--6214},
  year={2020}
}

@inproceedings{varol2021read,
  title={Read and Attend: Temporal Localisation in Sign Language Videos. In 2021 IEEE},
  author={Varol, Gul and Momeni, Liliane and Albanie, Samuel and Afouras, Triantafyllos and Zisserman, Andrew},
  booktitle={CVF Conference on Computer Vision and Pattern Recognition (CVPR)},
  pages={16852--16861},
  year={2021}
}

@inproceedings{ding2018audio,
  title={Audio-visual keyword spotting based on multidimensional convolutional neural network},
  author={Ding, Runwei and Pang, Cheng and Liu, Hong},
  booktitle={2018 25th IEEE international conference on image processing (ICIP)},
  pages={4138--4142},
  year={2018},
  organization={IEEE}
}

@inproceedings{stafylakis2018zero,
  title={Zero-shot keyword spotting for visual speech recognition in-the-wild},
  author={Stafylakis, Themos and Tzimiropoulos, Georgios},
  booktitle={Proceedings of the European Conference on Computer Vision (ECCV)},
  pages={513--529},
  year={2018}
}

@inproceedings{albanie2021seehear,
  title={SeeHear: Signer diarisation and a new dataset},
  author={Albanie, Samuel and Varol, G{\"u}l and Momeni, Liliane and Afouras, Triantafyllos and Brown, Andrew and Zhang, Chuhan and Coto, Ernesto and Camg{\"o}z, Necati Cihan and Saunders, Ben and Dutta, Abhishek and others},
  booktitle={ICASSP 2021-2021 IEEE International Conference on Acoustics, Speech and Signal Processing (ICASSP)},
  pages={2280--2284},
  year={2021},
  organization={IEEE}
}

@inproceedings{xu2024temporally,
  title={Temporally consistent unbalanced optimal transport for unsupervised action segmentation},
  author={Xu, Ming and Gould, Stephen},
  booktitle={Proceedings of the IEEE/CVF Conference on Computer Vision and Pattern Recognition},
  pages={14618--14627},
  year={2024}
}

@inproceedings{wang2022sscap,
  title={Sscap: Self-supervised co-occurrence action parsing for unsupervised temporal action segmentation},
  author={Wang, Zhe and Chen, Hao and Li, Xinyu and Liu, Chunhui and Xiong, Yuanjun and Tighe, Joseph and Fowlkes, Charless},
  booktitle={Proceedings of the IEEE/CVF Winter Conference on Applications of Computer Vision},
  pages={1819--1828},
  year={2022}
}

@inproceedings{tran2024permutation,
  title={Permutation-aware activity segmentation via unsupervised frame-to-segment alignment},
  author={Tran, Quoc-Huy and Mehmood, Ahmed and Ahmed, Muhammad and Naufil, Muhammad and Zafar, Anas and Konin, Andrey and Zia, Zeeshan},
  booktitle={Proceedings of the IEEE/CVF Winter Conference on Applications of Computer Vision},
  pages={6426--6436},
  year={2024}
}

@inproceedings{kumar2022unsupervised,
  title={Unsupervised action segmentation by joint representation learning and online clustering},
  author={Kumar, Sateesh and Haresh, Sanjay and Ahmed, Awais and Konin, Andrey and Zia, M Zeeshan and Tran, Quoc-Huy},
  booktitle={Proceedings of the IEEE/CVF Conference on Computer Vision and Pattern Recognition},
  pages={20174--20185},
  year={2022}
}

@inproceedings{shen2021learning,
  title={Learning to segment actions from visual and language instructions via differentiable weak sequence alignment},
  author={Shen, Yuhan and Wang, Lu and Elhamifar, Ehsan},
  booktitle={Proceedings of the IEEE/CVF Conference on Computer Vision and Pattern Recognition},
  pages={10156--10165},
  year={2021}
}

@article{yi2021asformer,
  title={Asformer: Transformer for action segmentation},
  author={Yi, Fangqiu and Wen, Hongyu and Jiang, Tingting},
  journal={arXiv preprint arXiv:2110.08568},
  year={2021}
}

@inproceedings{farha2019ms,
  title={Ms-tcn: Multi-stage temporal convolutional network for action segmentation},
  author={Farha, Yazan Abu and Gall, Jurgen},
  booktitle={Proceedings of the IEEE/CVF conference on computer vision and pattern recognition},
  pages={3575--3584},
  year={2019}
}

@article{li2020ms,
  title={Ms-tcn++: Multi-stage temporal convolutional network for action segmentation},
  author={Li, Shijie and Farha, Yazan Abu and Liu, Yun and Cheng, Ming-Ming and Gall, Juergen},
  journal={IEEE transactions on pattern analysis and machine intelligence},
  volume={45},
  number={6},
  pages={6647--6658},
  year={2020},
  publisher={IEEE}
}

@inproceedings{liu2023diffusion,
  title={Diffusion action segmentation},
  author={Liu, Daochang and Li, Qiyue and Dinh, Anh-Dung and Jiang, Tingting and Shah, Mubarak and Xu, Chang},
  booktitle={Proceedings of the IEEE/CVF international conference on computer vision},
  pages={10139--10149},
  year={2023}
}

@inproceedings{lu2024fact,
  title={Fact: Frame-action cross-attention temporal modeling for efficient action segmentation},
  author={Lu, Zijia and Elhamifar, Ehsan},
  booktitle={Proceedings of the IEEE/CVF Conference on Computer Vision and Pattern Recognition},
  pages={18175--18185},
  year={2024}
}

@inproceedings{shen2025understanding,
  title={Understanding Multi-Task Activities from Single-Task Videos},
  author={Shen, Yuhan and Elhamifar, Ehsan},
  booktitle={Proceedings of the Computer Vision and Pattern Recognition Conference},
  pages={19120--19131},
  year={2025}
}

@inproceedings{lu2025multi,
  title={Multi-modal few-shot temporal action segmentation},
  author={Lu, Zijia and Elhamifar, Ehsan},
  booktitle={Proceedings of the IEEE/CVF International Conference on Computer Vision},
  pages={14106--14116},
  year={2025}
}

@inproceedings{behrmann2022unified,
  title={Unified fully and timestamp supervised temporal action segmentation via sequence to sequence translation},
  author={Behrmann, Nadine and Golestaneh, S Alireza and Kolter, Zico and Gall, Juergen and Noroozi, Mehdi},
  booktitle={European conference on computer vision},
  pages={52--68},
  year={2022},
  organization={Springer}
}

@inproceedings{lee2024error,
  title={Error detection in egocentric procedural task videos},
  author={Lee, Shih-Po and Lu, Zijia and Zhang, Zekun and Hoai, Minh and Elhamifar, Ehsan},
  booktitle={Proceedings of the IEEE/CVF Conference on Computer Vision and Pattern Recognition},
  pages={18655--18666},
  year={2024}
}

@inproceedings{pang2024long,
  title={Long-tail temporal action segmentation with group-wise temporal logit adjustment},
  author={Pang, Zhanzhong and Sener, Fadime and Ramasubramanian, Shrinivas and Yao, Angela},
  booktitle={European Conference on Computer Vision},
  pages={320--338},
  year={2024},
  organization={Springer}
}

@inproceedings{shen2024progress,
  title={Progress-aware online action segmentation for egocentric procedural task videos},
  author={Shen, Yuhan and Elhamifar, Ehsan},
  booktitle={Proceedings of the IEEE/CVF Conference on Computer Vision and Pattern Recognition},
  pages={18186--18197},
  year={2024}
}

@inproceedings{wang2020boundary,
  title={Boundary-aware cascade networks for temporal action segmentation},
  author={Wang, Zhenzhi and Gao, Ziteng and Wang, Limin and Li, Zhifeng and Wu, Gangshan},
  booktitle={European Conference on Computer Vision},
  pages={34--51},
  year={2020},
  organization={Springer}
}

@inproceedings{ghoddoosian2023weakly,
  title={Weakly-supervised action segmentation and unseen error detection in anomalous instructional videos},
  author={Ghoddoosian, Reza and Dwivedi, Isht and Agarwal, Nakul and Dariush, Behzad},
  booktitle={Proceedings of the IEEE/CVF International Conference on Computer Vision},
  pages={10128--10138},
  year={2023}
}

@inproceedings{lu2022set,
  title={Set-supervised action learning in procedural task videos via pairwise order consistency},
  author={Lu, Zijia and Elhamifar, Ehsan},
  booktitle={Proceedings of the IEEE/CVF Conference on Computer Vision and Pattern Recognition},
  pages={19903--19913},
  year={2022}
}

@inproceedings{rahaman2022generalized,
  title={A generalized and robust framework for timestamp supervision in temporal action segmentation},
  author={Rahaman, Rahul and Singhania, Dipika and Thiery, Alexandre and Yao, Angela},
  booktitle={European Conference on Computer Vision},
  pages={279--296},
  year={2022},
  organization={Springer}
}

@inproceedings{sayed2023new,
  title={A new dataset and approach for timestamp supervised action segmentation using human object interaction},
  author={Sayed, Saif and Ghoddoosian, Reza and Trivedi, Bhaskar and Athitsos, Vassilis},
  booktitle={Proceedings of the IEEE/CVF Conference on Computer Vision and Pattern Recognition},
  pages={3133--3142},
  year={2023}
}

@inproceedings{shen2022semi,
  title={Semi-weakly-supervised learning of complex actions from instructional task videos},
  author={Shen, Yuhan and Elhamifar, Ehsan},
  booktitle={Proceedings of the IEEE/CVF Conference on Computer Vision and Pattern Recognition},
  pages={3344--3354},
  year={2022}
}

@inproceedings{fayyaz2020sct,
  title={Sct: Set constrained temporal transformer for set supervised action segmentation},
  author={Fayyaz, Mohsen and Gall, Jurgen},
  booktitle={Proceedings of the IEEE/CVF conference on computer vision and pattern recognition},
  pages={501--510},
  year={2020}
}

@inproceedings{singh2016multi,
  title={A multi-stream bi-directional recurrent neural network for fine-grained action detection},
  author={Singh, Bharat and Marks, Tim K and Jones, Michael and Tuzel, Oncel and Shao, Ming},
  booktitle={Proceedings of the IEEE conference on computer vision and pattern recognition},
  pages={1961--1970},
  year={2016}
}

@inproceedings{huang2020improving,
  title={Improving action segmentation via graph-based temporal reasoning},
  author={Huang, Yifei and Sugano, Yusuke and Sato, Yoichi},
  booktitle={Proceedings of the IEEE/CVF conference on computer vision and pattern recognition},
  pages={14024--14034},
  year={2020}
}

@article{zhang2022semantic2graph,
  title={Semantic2graph: graph-based multi-modal feature fusion for action segmentation in videos},
  author={Zhang, Junbin and Tsai, Pei-Hsuan and Tsai, Meng-Hsun},
  journal={arXiv preprint arXiv:2209.05653},
  year={2022}
}

@article{wang2024cross,
  title={Cross-enhancement transformer for action segmentation},
  author={Wang, Jiahui and Wang, Zhengyou and Zhuang, Shanna and Hao, Yaqian and Wang, Hui},
  journal={Multimedia Tools and Applications},
  volume={83},
  number={9},
  pages={25643--25656},
  year={2024},
  publisher={Springer}
}

@article{du2023dilated,
  title={Dilated transformer with feature aggregation module for action segmentation},
  author={Du, Zexing and Wang, Qing},
  journal={Neural Processing Letters},
  volume={55},
  number={5},
  pages={6181--6197},
  year={2023},
  publisher={Springer}
}

@article{xu2022don,
  title={Don't pour cereal into coffee: Differentiable temporal logic for temporal action segmentation},
  author={Xu, Ziwei and Rawat, Yogesh and Wong, Yongkang and Kankanhalli, Mohan S and Shah, Mubarak},
  journal={Advances in Neural Information Processing Systems},
  volume={35},
  pages={14890--14903},
  year={2022}
}

@inproceedings{duarte2021how2sign,
  title={How2sign: a large-scale multimodal dataset for continuous american sign language},
  author={Duarte, Amanda and Palaskar, Shruti and Ventura, Lucas and Ghadiyaram, Deepti and DeHaan, Kenneth and Metze, Florian and Torres, Jordi and Giro-i-Nieto, Xavier},
  booktitle={Proceedings of the IEEE/CVF conference on computer vision and pattern recognition},
  pages={2735--2744},
  year={2021}
}

@inproceedings{shi2022open,
  title={Open-domain sign language translation learned from online video},
  author={Shi, Bowen and Brentari, Diane and Shakhnarovich, Gregory and Livescu, Karen},
  booktitle={Proceedings of the 2022 Conference on Empirical Methods in Natural Language Processing},
  pages={6365--6379},
  year={2022}
}

@inproceedings{zhang2023c2st,
  title={C2st: Cross-modal contextualized sequence transduction for continuous sign language recognition},
  author={Zhang, Huaiwen and Guo, Zihang and Yang, Yang and Liu, Xin and Hu, De},
  booktitle={Proceedings of the IEEE/CVF International Conference on Computer Vision},
  pages={21053--21062},
  year={2023}
}

@inproceedings{wei2023improving,
  title={Improving continuous sign language recognition with cross-lingual signs},
  author={Wei, Fangyun and Chen, Yutong},
  booktitle={Proceedings of the IEEE/CVF International Conference on Computer Vision},
  pages={23612--23621},
  year={2023}
}

@inproceedings{gan2024signgraph,
  title={SignGraph: A Sign Sequence is Worth Graphs of Nodes},
  author={Gan, Shiwei and Yin, Yafeng and Jiang, Zhiwei and Wen, Hongkai and Xie, Lei and Lu, Sanglu},
  booktitle={Proceedings of the IEEE/CVF Conference on Computer Vision and Pattern Recognition},
  pages={13470--13479},
  year={2024}
}

@article{ye2024improving,
  title={Improving Gloss-free Sign Language Translation by Reducing Representation Density},
  author={Ye, Jinhui and Wang, Xing and Jiao, Wenxiang and Liang, Junwei and Xiong, Hui},
  journal={arXiv preprint arXiv:2405.14312},
  year={2024}
}

@inproceedings{yasser2024sign,
  title={Sign Language Translation with Sentence Embedding Supervision},
  author={Yasser, Hamidullah and Genabith, Josef and Espa{\~n}a-Bonet, Cristina},
  booktitle={Proceedings of the 62nd Annual Meeting of the Association for Computational Linguistics (Volume 2: Short Papers)},
  pages={425--434},
  year={2024}
}

@article{shen2024auslan,
  title={Auslan-daily: Australian sign language translation for daily communication and news},
  author={Shen, Xin and Yuan, Shaozu and Sheng, Hongwei and Du, Heming and Yu, Xin},
  journal={Advances in Neural Information Processing Systems},
  volume={36},
  year={2024}
}

@inproceedings{pavlakos2024reconstructing,
    title={Reconstructing Hands in 3{D} with Transformers},
    author={Pavlakos, Georgios and Shan, Dandan and Radosavovic, Ilija and Kanazawa, Angjoo and Fouhey, David and Malik, Jitendra},
    booktitle={CVPR},
    year={2024}
}

\end{document}